\documentclass{article}
\usepackage{spconf,amsmath,graphicx}

\title{Efficient adapter transfer of Self-Supervised speech models for Automatic Speech Recognition}
\name{Bethan Thomas$^\dagger$ \qquad Samuel Kessler\sthanks{Work performed during an internship with Huawei R\&D UK}$^\ddag$ \qquad Salah Karout$^\dagger$}

\address{$^\dagger$Huawei R\&D UK \qquad $^\ddag$University of Oxford}

\begin{document}
	%\ninept
	%
	\maketitle
	\begin{abstract}
		Self-supervised learning (SSL) is a powerful tool that allows learning of underlying representations from unlabeled data. Transformer based models such as wav2vec 2.0 and HuBERT are leading the field in the speech domain. Generally these models are fine-tuned on a small amount of labeled data for a downstream task such as Automatic Speech Recognition (ASR). This involves re-training the majority of the model for each task. Adapters are small lightweight modules which are commonly used in Natural Language Processing (NLP) to adapt pre-trained models to new tasks. In this paper we propose applying adapters to wav2vec 2.0 to reduce the number of parameters required for downstream ASR tasks, and increase scalability of the model to multiple tasks or languages. Using adapters we can perform ASR while training fewer than $10$\% of parameters per task compared to full fine-tuning with little degradation of performance. Ablations show that applying adapters into just the top few layers of the pre-trained network gives similar performance to full transfer, supporting the theory that higher pre-trained layers encode more phonemic information, and further optimizing efficiency.
	\end{abstract}

	\begin{keywords}
		Automatic Speech Recognition, Self-Supervision, Adapters, Transfer Learning
	\end{keywords}

	\section{Introduction}
	
	Self-supervision is now standard practice in Natural Language Processing (NLP), with models such as BERT \cite{bert} achieving state of the art results. More recently, self-supervised learning (SSL) approaches have been applied to speech tasks with great success. Recent works such as wav2vec 2.0 \cite{wav2vec2} and HuBERT \cite{hubert} show that self-supervision in speech can achieve state of the art results. 
	
	Most SSL models rely on unsupervised pre-training followed by supervised fine-tuning on a downstream task. The unsupervised pre-training allows a model to benefit from the large amounts of unlabeled data which are readily available. Fine-tuning uses a comparatively small amount of labeled data and retrains the majority of the self-supervised model to the desired downstream task.
	
	\begin{figure}[!t]
		
		\centering
		\centerline{\includegraphics[width=0.45\textwidth]{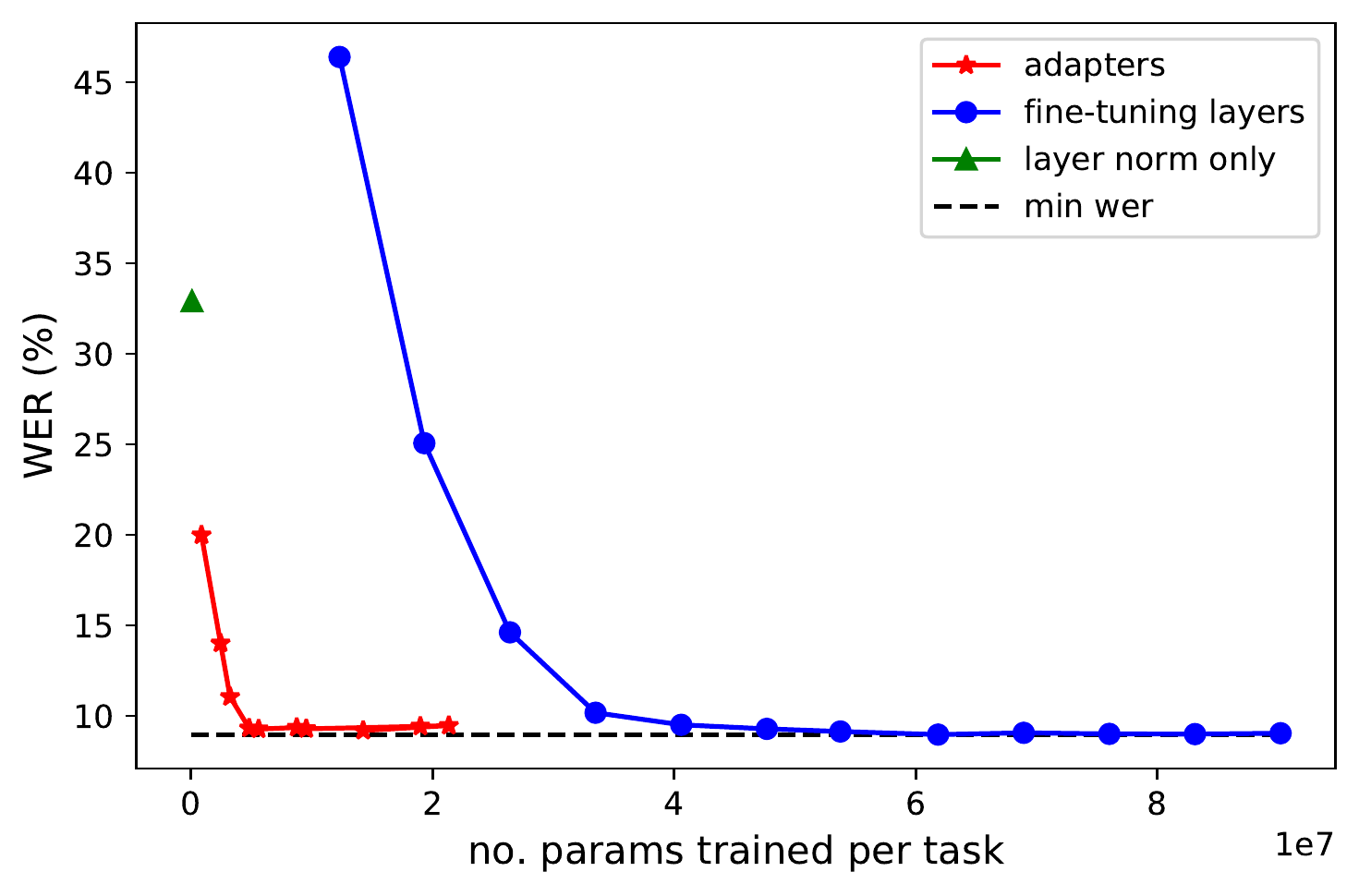}}
		
		\caption{\small{A comparison of trained model parameters vs WER. Pre-trained self-supervised models are either fine-tuned or adapted for ASR using 10 hours of supervised data and evaluated on LibriSpeech dev-clean. The Y axis shows various iterations of fine-tuning/adapters with different numbers of layers trained, and thus different numbers of trainable parameters. Adapters achieve a similar performance in WER to fine-tuning with only a fraction of parameters. Training just layer normalizations performs poorly. }}
		\label{fig:params}
	\end{figure}
	
	While these approaches gain excellent results, fine-tuning the model is computationally expensive and does not scale well to multiple tasks, such as in the case of multi-lingual Automatic Speech Recognition (ASR); a complete set of fine-tuned parameters must be learnt and stored per downstream task. This is commonly in the scale of $O(10^8)$ parameters per task. Once the model is fine-tuned for one task, the entire model is fixed for that task, and the base model must be reloaded to transfer to future tasks.
	
	Adapters are small trainable modules that can be applied into the layers of a frozen pre-trained network and tuned to a particular task. Recently these have been applied to pre-trained unsupervised models in NLP \cite{houlsby_adapter, madx_adpt}. The benefit of adapters is that they allow adaptation to new tasks with a relatively small number of parameters per task. As the existing model parameters remain frozen, the original model can support multiple downstream tasks, only a set of adapter parameters are required per task. This makes adapters more efficient to train, and highly scalable to multiple tasks.
	
	\begin{figure*}[!t]
		
		\centering
		\centerline{\includegraphics[width=0.95\textwidth]{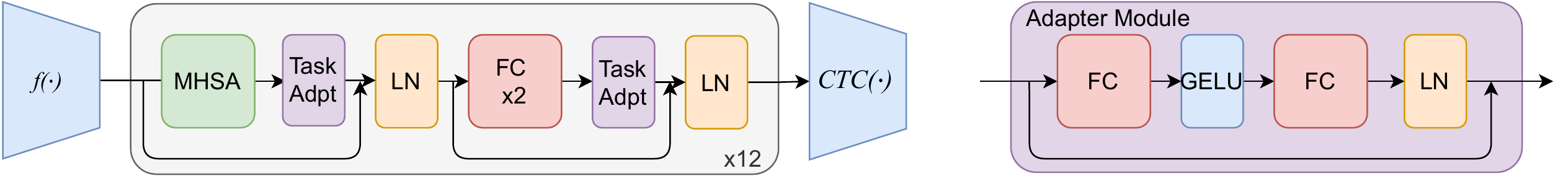}}
		
		\caption{\small{Left: structure of wav2vec 2.0 model with task specific adapter modules. \textit{f($\cdot$)} is a convolutional encoder which is followed by 12 transformer encoder blocks. For downstream ASR a linear classifier, \textit{CTC($\cdot$)}, is applied to the output of the transformer blocks. For adapter transfer, adapter modules are inserted into the model, and during adaptation to the downstream ASR task only the adapters, layer normalization layers and linear classifier are trained, the rest of the network is frozen. Right: structure of adapter module consisting of a down-projection, a non-linearity and an up-projection, with a skip connection.}}
		\label{fig:structure}
	\end{figure*}
	
	The task of ASR is very well researched for high resource languages such as English. However, it is more difficult to gain good performance on languages where there is limited paired text and audio training data. wav2vec 2.0 shows good performance with limited labeled data, and multi-lingual wav2vec 2.0 \cite{xlsr} shows good results across a range of languages including languages unseen during pre-training. Self-supervised speech models are able to leverage generic features from pre-training to adapt well to scenarios with limited labeled data and are thus ideal for multi-lingual ASR. However, it is impractical to train and store fine-tuned parameters for every language. Furthermore, fine-tuning overwrites pre-trained model parameters so may not be the optimal way of utilizing pre-training knowledge.

	In this paper we apply adapters to a pre-trained wav2vec 2.0 model in order to transfer the pre-trained representations to the task of downstream ASR. Adapters suggest a way of utilizing these pre-trained representations in a much more efficient manner, with many fewer parameters required per downstream task than usual fine-tuning methods. This has applications for multi-task scenarios such as multi-lingual ASR, and also increases the accessibility of self-supervised speech models for research. To the best of our knowledge this is the first time adapter modules have been applied to a self-supervised speech model for ASR.

	\section{Background}
	\subsection{Self-supervised ASR}
	Labeled speech data is relatively scarce and expensive to generate. Self-supervised methods are able to take advantage of the huge amount of unlabeled speech data available to learn generic speech features. 
	
	There have been various proposed methods such as CPC \cite{cpc}, APC \cite{apc}, and wav2vec \cite{wav2vec}. All aim to learn an encoded representation of a raw speech waveform using various self-supervised tasks which train the underlying network. Many of these models include some form of reconstruction, where masked representations, or elements of the original input waveform, are recreated. The underlying network can then be used to extract representations which are used as input to downstream tasks. This is instead of more conventional speech features such as log Mel filter-bank features. These networks can either be frozen, or further optimized to the downstream task by some form of transfer learning.
	
	wav2vec 2.0 is a leading model in this field \cite{wav2vec2}. However, the usual method of fine-tuning wav2vec 2.0 for downstream ASR requires re-training most of the model layers. In fact $95.6\%$ of the total model parameters are trained, and must therefore be stored, per task. This is not an issue if monolingual ASR is the only target task. However it is not ideal for multi-lingual speech recognition, or indeed utilizing the wav2vec 2.0 encoder for multiple tasks such as speech translation, as in \cite{speech_translation_adpt}. 
	
	Recently advances have been made to the wav2vec 2.0 architecture including adding self-training \cite{w2v_self}. While we use wav2vec 2.0 in this work, our approach would scale well to any self-supervised transformer based speech model, including HuBERT \cite{hubert}.

	\subsection{Adapters}
	
	Adapters were introduced in NLP as an efficient method of transfer learning \cite{houlsby_adapter}. It was found that adapters approach the performance of full fine-tuning with only a fraction of the parameters in NLP tasks using a pre-trained BERT model. Adapters have also been successfully applied to NMT \cite{bapna_adapter} and vision tasks \cite{vision_adapter}. Recently dual adapters have been proposed for multi-lingual transfer \cite{madx_adpt}, where one adapter module is used to adapt for language, and another captures task specific adaptations.
	
	Adapters have also been applied to speech in streaming RNNT models \cite{speech_adapters1} and hybrid CTC-attention speech transformers \cite{speech_adapters_dual} to solve the issue of multi-lingual speech recognition. More recently adapters have been used in speech translation to combine pre-trained modules \cite{speech_translation_adpt}. Adapters have also been employed for efficient SSL pre-training of new tasks in a continual learning setting \cite{kessler2021continual}. 
	
	We hypothesize that we will see the same benefits of adapters in a speech model as in an NLP model, namely parameter efficient transfer of the pre-trained network to a downstream task with little performance degradation.

	\section{Method}\label{method}

	In wav2vec 2.0 a convolutional encoder takes as input the raw waveform. Then a transformer model is applied to the encoded output. The model is trained by masking input frames and comparing predicted values with positive and negative quantized versions of the masked frames. This is similar to contrastive predictive coding \cite{cpc}.
	
	When applying wav2vec 2.0 to a downstream model the feature encoder is frozen, and a linear classifier is added on top of the transformer context model for fine-tuning. Generally the transformer model is also frozen for the first N updates. When applying wav2vec 2.0 to ASR, the linear classifier and model are fine-tuned with a Connectionist Temporal Classification (CTC) loss using a small amount of labeled speech data. 
	
	Adapters are small bottleneck modules consisting of a down projection, a non-linearity, and an up-projection, with a skip connection (see Fig. \ref{fig:structure}). The initial implementation \cite{houlsby_adapter} applies adapters after both the self-attention and feedforward layers. However it is possible to apply adapters in different positions throughout the transformer block \cite{bapna_adapter}. The fully connected layers are initialized as a near identity function. The identity initialization and the skip connection allow the module to be ignored if not deemed necessary during training. 
	
	In our experiments with adapters we apply adapters twice in each transformer block. Our experimentation showed that this gave the best results. We apply a linear classifier on top of the transformer network. A set of adapters, layer normalization layers and a linear classifier are trained per task using a CTC loss, the rest of the network remains frozen.

	\section{Experiments and Results}

	We use the standard wav2vec 2.0 BASE architecture which contains 12 transformer layers, and use the publicly released pre-trained checkpoint from \texttt{fairseq} \cite{fairseq} which has been pre-trained using the LibriSpeech \cite{librispeech} dataset. We use a standard size of 256 for all adapters, and initially apply adapters into every transformer layer.
	
	We also run our own fine-tuning experiments as a comparison and follow the fine-tuning format of wav2vec 2.0. By tuning hyper-parameters we are able to improve on the word error rate (WER) values reported in that paper \cite{wav2vec2}. We trained for 20k steps, with the transformer layers frozen for just the first 4k updates, and a learning rate of 5e-5.

	\begin{table}[h!]
		\caption{\small{A pre-trained BASE wav2vec2 model is transferred to the downstream ASR task using the 10 hour LibriLight (LL) supervised set, or a random 10 hour subset of the French Common Voice (CV) corpus. Word error rate (WER) is reported on the dev-clean/dev-other sets of LibriSpeech for English, and the CV test set for French. Results are all without a language model. \% of trainable parameters compared to the total model parameters are also reported}}
		\centering
		\scalebox{0.9}{
		\begin{tabular}{lccc} \hline\hline
			&wav2vec 2.0 FT\cite{wav2vec2} 		&Fine-tune 		&Adapter		\\
			\hline
			10h LL dev-clean	&$10.9\%$ &$8.98\%$  &$9.39\%$ \\
			10h LL dev-other  	&$17.4\%$ &$16.9\%$ &$17.0\%$ \\
			\hline
			10h French CV test			&N/A	&$40.2\%$ 	&$39.4\%$	 \\
			\hline\hline
			\% trained params	&$95.6\%$	&$95.6\%$	&$9.2\%$ \\
			\hline\hline
		\end{tabular}
	}
		%}
		\label{tab:ASR}
	\end{table}

	We investigate performance of fine-tuning and adapters on the 10 hour supervised subset of the LibriLight (LL) dataset \cite{librilight}, and evaluate on standard LibriSpeech dev sets. We also experiment with French ASR to demonstrate the multi-task scenario. We take a 10h subset of the Common Voice (CV) \cite{commonvoice} corpus and evaluate on the CV test set. We calculate WER in all cases. We found that the optimal setup for adapter transfer was to run for 10k steps with a learning rate of 5e-4. All experiments are run on 8 V100 GPUs, and without language model fusion to enable distraction free comparison of transfer method.
	
	%\begin{table}[!h]
	%	\caption{WER on dev-clean/dev-other sets of LibriSpeech. A pre-trained BASE wav2vec2 model is transferred to the downstream ASR task using the 10hour LibriLight supervised set. Results are all without a language model} 
	%	\begin{center}
	%		\begin{tabular}{cccc} \hline\hline
	%			10h LL			&Dev-clean 		&Dev-other 		&No. trained		\\
	%							&WER			&WER			&params 			\\
	%			\hline
	%			W2v2 ft \cite{wav2vec2} 	&$10.9\%$	&$17.4\%$   &$95.6\%$   			\\
	%			\hline
	%			Fine-tune (own) &$8.98\%$		&$16.86\%$		&$95.6\%$				\\
	%			Adapter (own) 	&$9.39\%$		&$17.03\%$		&$9.2\%$	\\
	%			\hline\hline
	%		\end{tabular}
	%	\end{center}
	%	%}
	%	\label{tab:ASR}
	%\end{table}
	%
	%
	%\begin{table}[!h]
	%	\caption{PER results on Common Voice French test set for a base English wav2vec 2.0 model fine-tuned with 10 hours and 1 hour of data French data.} 
	%	\begin{center}
	%		\begin{tabular}{cccc} \hline\hline
	%						&10h 		&1h 		&No. trained		\\
	%						&PER		&PER		&params 			\\
	%			\hline
	%			XLSR ft \cite{xlsr} 	&N/A	&$20.0\%$   &$95.6\%$   			\\
	%			\hline
	%			Fine-tune (own) &$10.7\%$		&$19.4\%$		&$95.6\%$				\\
	%			Adapter (own) 	&$11.1\%$		&$20.7\%$		&$9.2\%$	\\
	%			\hline\hline
	%		\end{tabular}
	%	\end{center}
	%	%}
	%	\label{tab:ASR}
	%\end{table}
	
	\begin{figure}[t]
		
		\centering
		\centerline{\includegraphics[width=0.45\textwidth]{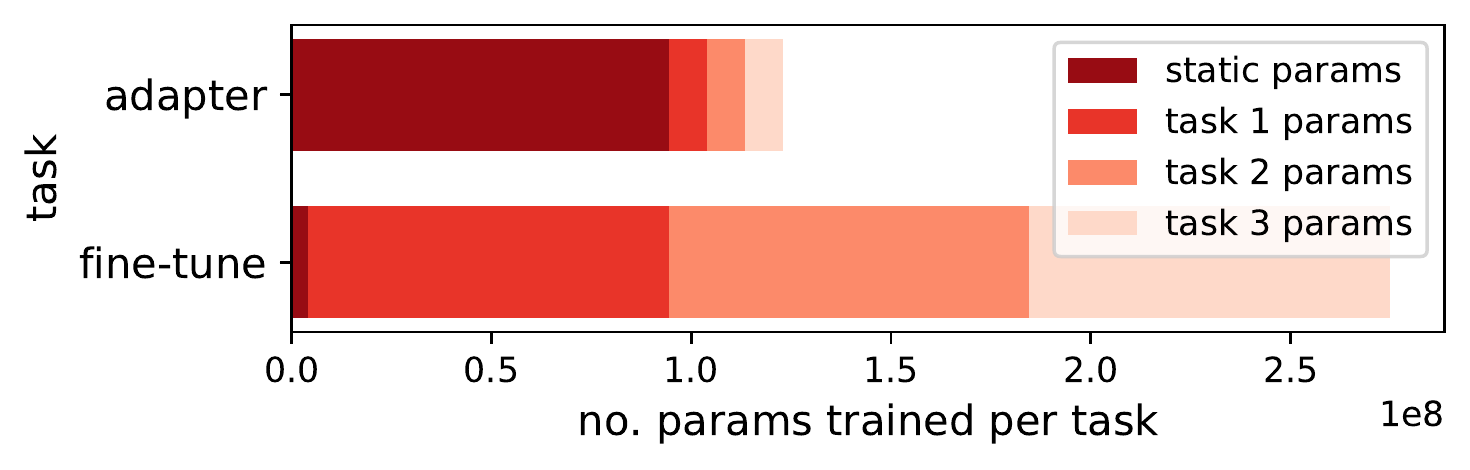}}
		
		\caption{\small{A summary of number of trained parameters per task with fine-tuning and adapters. Each fine-tuning task nearly doubles the number of trained parameters which must be learnt and stored. Whereas adapters add only a small number of additional parameters per task showing their scalability to multi-task scenarios.}}
		\label{fig:tasks}
		\vspace{-0.5cm}
	\end{figure}

	Adapters perform slightly worse than fine-tuning on English ASR (see table \ref{tab:ASR}), however the absolute WER increase is just $0.41\%$ and $0.17\%$ for dev-clean and dev-other respectively. This performance decrease comes with a huge decrease in the number of trained task-specific parameters. Fine-tuning requires training $95.6\%$ of parameters, our adapter approach only trains $9.2\%$ of parameters. For French ASR, there is in fact a slight performance increase when using adapters compared to traditional fine-tuning.
	
	However, the real benefit of this approach comes when scaling to multiple languages or tasks, as can be seen in Fig. \ref{fig:tasks}. Each fine-tuning task nearly doubles the required number of parameters which must be learnt and stored, however adapters add only a small number of additional parameters per task. This makes adapter transfer much more scalable than fine-tuning while attaining a similar performance.  
	
	Our results show that even when there is a mismatch between pre-trained and downstream language, in the case of French transfer, both fine-tuning and adapters are able to achieve some success in ASR. This is unsurprising for fine-tuning, since the language specific features of the pre-trained network are overwritten during training. However, with adapter transfer, the original network remains unchanged. Therefore the adapters are able to compensate for the language mismatch, and learn both language and task specific features. This is in contrast to \cite{madx_adpt} which uses separate adapters for task and language.
	
	It is also worth noting that adapter experiments run more quickly than fine-tuning experiments, both due to the reduction in trainable parameters which increases training speed, and the smaller optimal number of training steps. Adapter experiments do not depend on a freeze steps hyper-parameter, and run successfully with a simple bi-stage learning rate schedule, as in \cite{houlsby_adapter}, rather than the more complex tri-stage scheduler required for traditional fine-tuning \cite{wav2vec2}. All these factors make this adapter approach much more experimentally friendly for researchers.

	\begin{table}[!t]
		
		\caption{\small{A pre-trained bilingual (French and English) model is transferred to English and French downstream ASR using fine-tuning and adapters. Results are reported in WER.}}
		\centering
		\begin{tabular}{lcc} \hline\hline
			&Fine-tune 		&Adapter		\\
			\hline
			10h LL dev-clean	 &$12.16\%$  &$12.91\%$ \\
			\hline
			10h French CV test		&$27.75\%$ 	&$28.25\%$	 \\
			\hline\hline
		\end{tabular}
		%}
		\label{tab:french_ASR}
		\vspace{-0.5cm}
	\end{table}
	
	We also pre-trained our own bi-lingual English and French wav2vec 2.0 model using approximately 1000 hours of French CV data, as well as 960h of English LibriSpeech data, and ran English and French ASR experiments (see table \ref{tab:french_ASR}). This demonstrates that our adapter method is also valid for multi-lingual pre-trained models as adapters again get within close performance of fine-tuning.
	
	Finally, we tested training just the layer normalization parameters as suggested in \cite{houlsby_adapter}. However, as found in that work, this approach does not perform well, WER on LibriSpeech dev-clean was $32.9\%$. This provides evidence that the adapter modules themselves improve performance, rather than the re-trained layer normalization parameters.

	\subsection{Ablations}

	Prior work suggests that lower layers of self-supervised models contain more generic speech features, and higher layers contribute more to phone discrimination \cite{apc}. Therefore we investigate training just the top N layers of the network using both fine-tuning and adapters. All experiments are run with the BASE English wav2vec 2.0 model, 10 hours LibriLight training data and evaluated on LibriSpeech dev-clean.
	
\begin{figure}[!t]
		
		\centering
		\centerline{\includegraphics[width=0.45\textwidth]{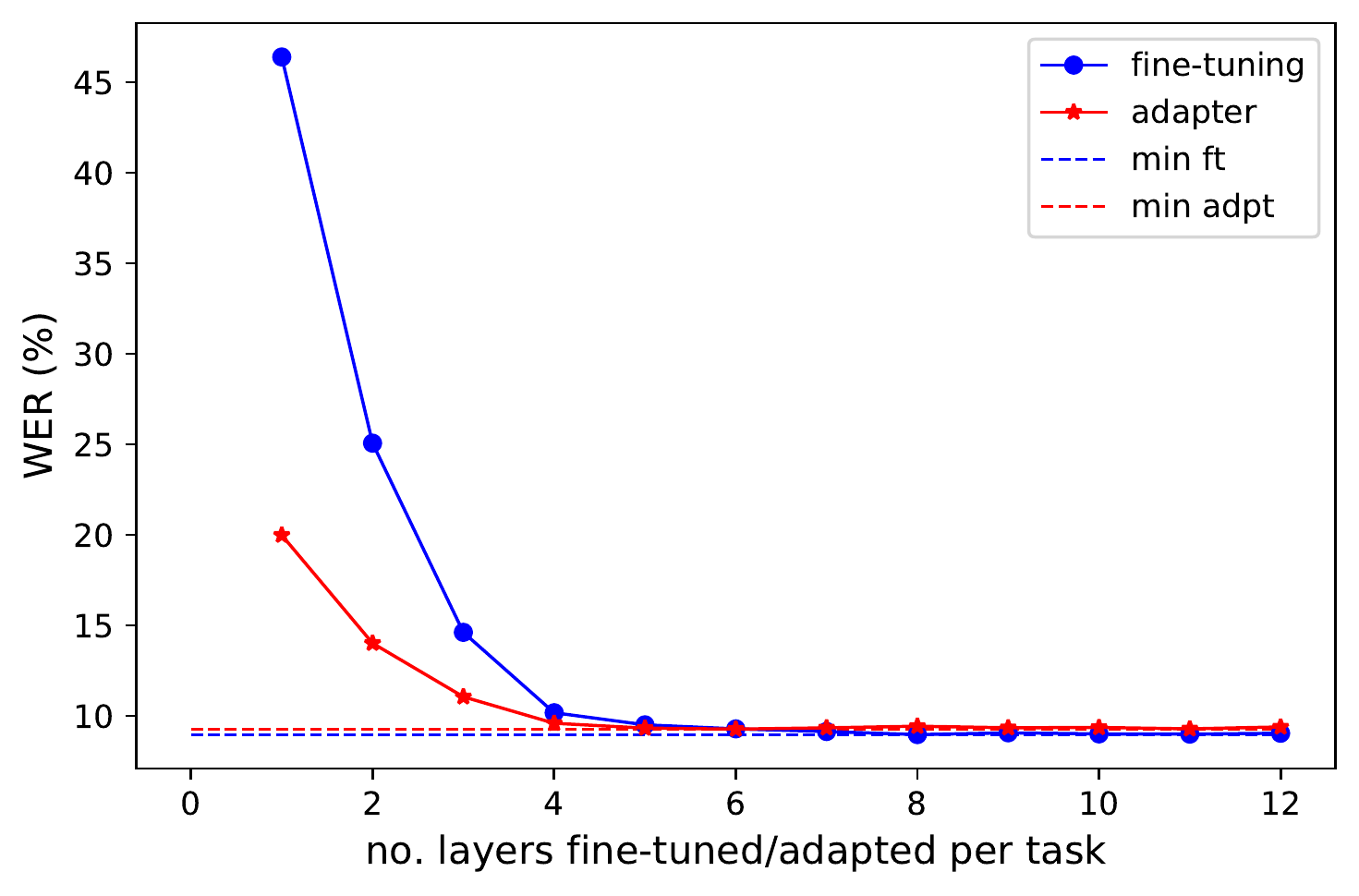}}
		
		\caption{\small{A comparison of trainable layers vs WER. For fine-tuning experiments, transformer layers of the model are optionally trained or frozen in a top down manner and evaluated for WER on LibriSpeech dev-clean. For adapter experiments, adapters are only inserted into the top N layers. Models with adapters perform better than models which are fine-tuned when the number of trained layers is limited. }}
		\label{fig:ablations}
		\vspace{-0.5cm}
	\end{figure}

	Using adapters in just the top 4 layers, out of a total 12 layers, gives nearly as good performance as adding adapters into every layer, and in fact using just 6 adapters gives the best performance with $9.27\%$ WER. This immediately halves the number of required parameters to just $4.85$\% of total parameters trained. It is also possible to optimize fine-tuning; training just the top 8 transformer layers gives best performance. However fine-tuning requires more layers to be trained than using adapters, and even 8 layers equates to training $65.5\%$ of parameters (see Fig. \ref{fig:ablations}). When just one layer is trained, the method with adapters performs significantly better showing that adapters are better able to utilize the pre-training knowledge of the entire network. 
	
	The curves presented in Fig. \ref{fig:ablations} show that the top layers of the network are more important for downstream performance than the bottom layers, supporting the hypothesis that the higher layers of the network encode more phonemic information, and lower layers more generic speech information.
	
	\section{Discussion \& Conclusion}
	This work provides an insight into how self-supervised speech models can be utilized in a more parameter efficient manner without sacrificing performance. When transferring to a downstream task, fine-tuning the majority of the model is still generally the best performing method. However, using adapters instead of fine-tuning achieves close to that performance with a fraction of parameters and takes less training time. This allows quicker and cheaper experimentation with models such as wav2vec 2.0 and HuBERT, and increases scalability to multiple tasks, for example multi-lingual ASR. More work could be done on utilizing adapters for additional tasks such as speaker recognition and speech translation.
	
	Ablations show that it is unnecessary to transfer all layers of the network to the downstream task, only the top N layers, thereby supporting the theory that the higher layers of these pre-trained networks encode more phonemic information. Adapters are better able to utilize pre-trained information as we achieve better performance with one layer of adapters, than with one layer of fine-tuning, and best performance comes from adapting fewer layers compared to fine-tuning. 
	
	While we are the first to utilize adapters in this way for speech, our findings are similar to those in the NLP domain \cite{houlsby_adapter, madx_adpt}. It would be interesting to perform the same layer ablations in the NLP domain. More broadly, we show that the speech domain can benefit from future NLP work on SSL.

	\bibliographystyle{IEEEbib}
	\bibliography{refs}
	
\end{document}